# HazeSpace2M: A Dataset for Haze Aware Single Image Dehazing

Md Tanvir Islam
Sungkyunkwan University
Suwon, Republic of Korea
tanvirnwu@g.skku.edu

Nasir Rahim
Sungkyunkwan University
Suwon, Republic of Korea
nrahim3797@skku.edu

Saeed Anwar
The Australian National University
Canberra, Australia
saeed.anwar@anu.edu.au

Muhammad Saqib
National Collections & Marine
Infrastructure, CSIRO, Marsfield
Sydney, Australia
muhammad.saqib@data61.csiro.au

Sambit Bakshi
National Institute of Technology
Rourkela, India
bakshisambit@ieee.org

Khan Muhammad*
Department of Human-AI Interaction
Sungkyunkwan University
Seoul, Republic of Korea
khanmuhammad@g.skku.edu

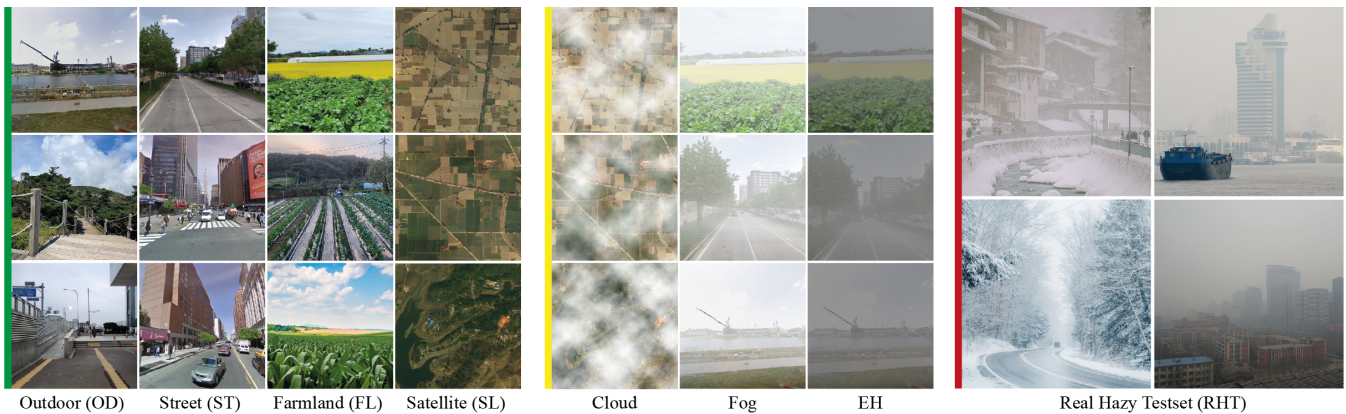

Figure 1: Ground Truth (Green), Synthetic Hazy (Yellow), and Real Hazy (Red) images across various scenes of HazeSpace2M.

## ABSTRACT

Reducing the atmospheric haze and enhancing image clarity is crucial for computer vision applications. The lack of real-life hazy ground truth images necessitates synthetic datasets, which often lack diverse haze types, impeding effective haze type classification and dehazing algorithm selection. This research introduces the HazeSpace2M dataset, a collection of over 2 million images designed to enhance dehazing through haze type classification. HazeSpace2M includes diverse scenes with 10 haze intensity levels, featuring Fog, Cloud, and Environmental Haze (EH). Using the dataset, we introduce a technique of haze type classification followed by specialized dehazers to clear hazy images. Unlike conventional methods, our approach classifies haze types before applying type-specific dehazing, improving clarity in real-life hazy images. Benchmarking with state-of-the-art (SOTA) models, ResNet50 and AlexNet achieve 92.75% and 92.50% accuracy, respectively, against existing synthetic datasets. However, these models achieve only 80% and 70% accuracy, respectively, against our Real Hazy Testset (RHT), highlighting the challenging nature of our HazeSpace2M dataset. Additional experiments show that haze type classification followed by specialized dehazing improves results by 2.41% in PSNR, 17.14% in SSIM, and 10.2% in MSE over general dehazers. Moreover, when testing with SOTA dehazing models, we found that applying our proposed framework significantly improves their performance. These results underscore the significance of HazeSpace2M and our proposed framework in addressing atmospheric haze in multimedia processing. Complete code and dataset is available on [GitHub](#).

## CCS CONCEPTS

• **Computing methodologies** → **Computer vision tasks**.

## KEYWORDS

HazeSpace2M, Haze type classification, Haze aware dehazing, Single image dehazing, Haze classification, Atmospheric haze

*Corresponding author

## 1 INTRODUCTION

Atmospheric haze significantly compromises image clarity, posing difficulties for computer vision tasks in autonomous systems, remote sensing, and surveillance [70]. Adverse weather conditions that reduce visibility can lead to accidents, as documented in various studies [23, 32, 43, 61]. To tackle the issues of hazes, researchers have developed dehazing algorithms to counteract haze's effects on image quality [7]. Advancements in traffic systems and vehicle detection technology further necessitate enhanced visibility [15, 43, 57, 71]. Current efforts focus on refining models to restore clarity to images impaired by adverse environmental conditions [44, 63, 65]. However, there is a consensus on the necessity for versatile dehazing techniques across variable weather patterns [20], with a rich dataset being crucial for developing robust dehazing models for effective atmospheric dehazing. Because emerging fields, e.g., agriculture and environmental studies, benefit from sophisticated image processing across different domains using Unmanned Aerial Vehicles (UAVs) to gather diverse scenes, including aerial views [48]. These applications require analyzing high-resolution images often obscured by various haze types, such as Fog and EH in outdoor, street, and farmland settings and cloud in aerial views.

Large datasets with varied scenes and haze types are scarce; the RESIDE SOTS [34] benchmark dataset covers synthetic hazy images but is limited to a single haze type. Similarly, the Cityscapes [12] dataset includes fog and rain conditions but is confined to street scenes, highlighting a deficit in comprehensive hazy image datasets. Current image restoration (IR) models often operate without recognizing the specific degradation type [33, 39, 45, 46, 62, 66, 69]. Although instruction-based IR methods [8, 11] improve performance by classifying degradation type, they rely on manual input, which is impractical for autonomous systems. An automated model that can identify and adapt to various haze types in the image is needed for effective dehazing without human intervention.

However, to train such versatile models, we need a dataset that offers various haze types across different scene types [20], which is absent in the literature. Identifying this gap in the literature, we develop a dataset named "HazeSpace2M," suitable for haze type classification and training haze type-specific specialized dehazers. We structure the "HazeSpace2M" dataset in a way that is suitable for haze type classification and training haze type-specific specialized dehazers. Leveraging this dataset in this paper, we also propose a novel idea of an intelligent image dehazing approach that performs specialized dehazing based on the haze type present in an input image. Thus, our research makes significant progress in the direction of image dehazing and classification, marked by the following contributions:

- `Development of a Benchmarking Dataset:` We developed HazeSpace2M as a comprehensive benchmarking dataset designed explicitly for haze type classification. Additionally, we are the first to introduce a hazy dataset for different scene types, especially the Farmland scene type, which is unparalleled in the literature. This dataset surpasses existing datasets in terms of number of images (over 2 million), scene types, type of hazes, and haze intensity (10 levels).

- `Intelligent Haze Aware Dehazing:` We propose a novel framework that performs dehazing with specialized dehazers based on the haze type present in the input hazy image.
- `Benchmarking SOTA Models:` We evaluate leading classification and dehazing models, setting new benchmarks for haze type classification and dehazing.
- `Evidence for Specialized Dehazers' Efficacy:` Our findings show that specialized dehazers, informed by accurate haze type classification, enhance dehazing performance, surpassing the capabilities of generalized dehazing models.

## 2 RELATED WORKS

In recent years, various hazy image datasets [4, 5, 12, 25, 34, 37, 50, 58, 59] have emerged to aid in developing single image dehazing models. These datasets offer a range of images affected by different hazes. For instance, the RESIDE dataset [34] compiles a variety of images, including both indoor and outdoor settings, with hazy conditions and their corresponding ground truth (GT) images. However, it lacks distinct subsets for diverse images and haze conditions. Conversely, the Cityscapes dataset [12] provides fog and rain-afflicted street scenes but lacks variety in scene types. The synthetic image collections FRIDA [59] and FRIDA2 [58] are designed primarily for algorithm assessment in visibility and contrast restoration, encompassing 90 and 330 images across urban road scenes, respectively. Despite their utility, the synthetic nature of these sets limits their effectiveness in modeling the complexity of real-world hazes.

To bridge the above mentioned gaps, the NH Haze [6] dataset, introduced during the NTIRE2020 [1] challenge, features 55 outdoor scenes with actual haze conditions alongside their haze-free GT images, proving invaluable for developing new dehazing methods. Moreover, the Haze4k dataset [37]-split into 3,000 training and 1,000 testing images—provides ample data for benchmarking novel dehazing approaches. Adding diversity, the Kede [40] dataset contains 225 images with nine groups showcasing different outdoor settings and haze thicknesses. In contrast, the O-HAZE [5] dataset with 45 scenes captured under consistent lighting conditions offers realistic pairs of hazy and clear images, facilitating the study of dehazing in authentic environments. In the realm of remote sensing, datasets like Haze1k [25] and RS Haze [50] enrich the dehazing research by providing images categorized by haze density and showcasing a variety of cloud haze levels, respectively. Haze1k offers 900 images curated for remote sensing applications, whereas RS Haze challenges researchers with nine distinct haze levels in its 5,700 GT images. These datasets play a crucial role in enhancing the development of algorithms that deal with the nuances of hazy conditions observed in satellite imagery.

Overall, these datasets have become central to benchmarking the performance of single image dehazing techniques. Especially, the datasets like RESIDE [34] and Foggy Cityscapes [12] with their extensive collection, are excellent for generalizing models and have become a benchmark for assessing dehazing algorithms [9, 16–18, 24, 28, 29, 36, 42, 50, 52]. However, the ranges of haze types and scenes are limited, scoping the improvement with more diverse datasets having various haze types for creating classification models capable of classifying various haze conditions. To fill this gap, we present a new dataset that is both broad and diverse, covering a wide range of scene types and haze types, paving the way for breakthroughs in the realm of single image dehazing in terms of developing robust haze type classification and dehazing algorithms.

Table 1: Overview of HazeSpace2M dataset scene and haze types, each with 10 distinct haze intensity levels.

| HazeSpace2M | | | |
|---|---|---|---|
| Outdoor | Street | Farmland | Satellite |
| Fog | Fog | Fog | Cloud |
| EH | EH | EH | |
| 10 different levels of haze for each category | | | |

## 3 OUR DATASET: HAZESPACE2M

HazeSpace2M is a diverse and large dataset with over 2M images, including the GT and Hazy images of three different types of hazes: Fog, Environmental Haze (EH), and Cloud. To the best of our knowledge, we are the first to introduce both EH and Fog separately for scenarios such as Outdoor, Street, and Farmlands. HazeSpace2M is suitable for developing intelligent dehazing models based on haze type classification. Notably, the HazeSpace2M includes four main scene categories: Outdoor, Street, Farmland, and Satellite, encompassing three haze conditions: Fog, EH, and Cloud, as stated in Table 1. Each GT image from every scene type features ten corresponding hazy images, varying from low to high intense levels. The HazeSpace2M dataset includes an extensive collection of over 130,193 GT images and approximately two million hazy images, each categorized into distinct levels of haze intensity across various scene types. It also has a subset named Real Hazy Testset (RHT) that features 1,030 real hazy images for evaluating models. This comprehensive dataset not only paves the way for creating more robust dehazing models but also facilitates the development of algorithms capable of classifying the types of haze present, thereby contributing significantly to image processing and multimedia.

### 3.1 Data Collection and Generation

We collected a large amount of image data from various sources to serve as the GT images in the HazeSpace2M dataset.

**Quality Assurance.** As shown in Table 2, we collect most of our GT images from the existing datasets or online under a Creative Commons License (CML), and some are the images captured from our personal devices. Cross-checking the quality of these images is a challenging but essential task. Initially, to ensure the quality of the GT images of our HazeSpace2M dataset, we established three conditions for excluding GT images as follows:

- `Resolution`: The image is of low quality.
- `Haze Presence`: The image contains haze in any form.
- `Irrelevance`: The image is not relevant to the scene types of HazeSpace2M.

If any image from our sources meets either of these criteria, it is excluded from the GT set of the HazeSpace2M dataset. For example, as shown in Table 2, we selected only 2,106 images from the ADE20k [73, 74] dataset out of 27,638 and 7,851 images out of 8,964 from the RESIDE SOTS [34] to use as GT images in the HazeSpace2M dataset. Similarly, we take 20,000 out of 62,068 images from the GSV [68] dataset as the rest match criteria 3. Thus, we ensure the quality and reusability of the GT images while we collect the GT images for HazeSpace2M from a wide range of sources.

Table 2: A breakdown of the various image sources and the number of GT images selected from each source.

| Scene Types | Image Sources | Total # of Images in Source | Total # of GT Images we Pick |
|---|---|---|---|
| Outdoor (OD) | ADE20K [73, 74] | 27,638 | 2,106 |
| | OTS [34] | 8,964 | 7,851 |
| | GSV [68] | 62,068 | 20,696 |
| | SFTGAN [64] | 10,200 | 4,596 |
| | Our Collections | 687 | 687 |
| Street (ST) | GSV [68] | 62,068 | 20,000 |
| | Cityscapes [12] | 19,998 | 19,998 |
| Farmland (FL) | Our Collections | 830 | 830 |
| Satellite (SL) | Haze1k [25] | 1,035 | 898 |
| | Forest Fires [19] | 42,815 | 42,815 |
| | DGLCC [13] | 1,146 | 1,146 |
| | DGRED [13] | 8,570 | 8,570 |
| **Total GT Images:** | | | **130,193** |

**Scene Types.** As mentioned earlier, our HazeSpace2M dataset comprises diverse scenes. Outdoor images provide aerial and ground-level views of urban environments, capturing elements like architecture and traffic. Street view offers a closer look at urban roads and daily life. Farmland images focus on agricultural areas, detailing rural landscapes. Satellite images from high altitudes afford expansive views of the Earth's valuable surface for geographical and environmental studies and tracking changes in land use patterns, highlighting details unnoticeable at ground level. The images with the green line in Figure 1 display some sample images of different scenes of the HazeSpace2M dataset.

**Haze Types.** The HazeSpace2M dataset features three hazes types: Fog, EH, and Cloud. The haze types are applied to the GT images to create ten different haze intensities, which means that from each GT image, we produce ten hazy images of different haze intensity, which varies from light to dense.

*Fog:* Fog is caused by the presence of water droplets in the air, typically when there is a high relative humidity. It is a ground-level haze that reduces visibility.

*Cloud:* Cloud haze is characterized by cloud formations at various altitudes, affecting the lighting and contrast in images.

*Environmental Haze (EH):* EH is an atmospheric condition characterized by fine particles, aerosols, and pollutants suspended in the air. It is commonly caused by human activities, including industrial emissions and vehicle exhaust, but can also originate from natural sources such as burning from wildfires and agricultural lands. The images with the yellow line in Figure 1 display some sample images of different haze types of the HazeSpace2M dataset.

**Real Haze Testset (RHT):** The RHT comprises a collection of real-life hazy images sourced online to evaluate the ability of our classification models to identify haze types in real-world scenarios. These images are curated using specific search terms; for instance, searches for "foggy images," "foggy weather," and "winter fog" helped label images as Fog. Similarly, searches using "environmental haze," "air pollution," "wildfire," and "smoky environment" facilitated the labeling of images as EH. We meticulously verify each image's visual characteristics and origin to accurately represent the specified haze type. Thus, we collected around 686 images with fog haze and 344 with EH. The images with the red line in Figure 1 present some sample images of the RHT subset of the HazeSpace2M dataset. However, sourcing original satellite images depicting cloud haze posed a challenge. To address this, we incorporated 500 satellite cloudy images from the RS Haze [50] dataset into RHT, enabling comprehensive evaluation of the classifiers.

Table 3: Number of GT and hazy images in HazeSpace2M subsets by scene type and haze condition, with subset names.

| Subset Names of HazeSpace2M depending on various Scene Types | Subsets Names of HazeSpace2M depending on various Haze Types | HazeSpace2M | | |
|---|---|---|---|---|
| | | Nature of the Image | # of GT Images | # of Hazy Images |
| Outdoor (OD) | Outdoor Fog (ODF) | Synthetic | 35,936 | 359,360 |
| | Outdoor Environmental Haze (ODEH) | Synthetic | | 359,360 |
| Street (ST) | Street Fog (STF) | Synthetic | 39,998 | 399,980 |
| | Street Envirnmental Haze (STEH) | Synthetic | | 399,980 |
| Farmlands (FL) | Farmland Fog (FLF) | Synthetic | 830 | 8,300 |
| | Farmland Envirnmental Haze (FLEH) | Synthetic | | 8,300 |
| Satellite (SL) | Satellite Cloud (SLC) | Satellite | 53,429 | 534,290 |
| **Real Haze Testset (RHT)** | - | Real | - | 1,030 |
| | Total: | | 130,193 | 2,070,600 |
| | Total # of Images (GT + Hazy) in HazeSpace2M dataset: | | 2,200,793 (2.2 Million Images) | |

Table 4: Comparative evaluation of image quality metrics across the existing datasets. The comparison of PSNR and SSIM values for the lowest and highest haze levels across different datasets, including our HazeSpace2M dataset.

| Datasets | Scene Type | Haze types | | | # of GT Images | # of Hazy Images | Lowest Haze Level | | Highest Haze Level | |
|---|---|---|---|---|---|---|---|---|---|---|
| | | Fog | Cloud | EH | | | PSNR ↑ | SSIM ↑ | PSNR ↑ | SSIM ↑ |
| FRIDA [58, 59] | Outdoor | ✓ | ✗ | ✗ | 84 | 420 | 27.54 | 0.81 | 29.92 | 0.69 |
| Foggy Driving [12] | Street | ✓ | ✗ | ✗ | 10,425 | 10,425 | 28.50 | 0.88 | 27.70 | 0.68 |
| I-Haze [37] | Outdoor | | Not Specified | | 30 | 30 | 29.34 | 0.85 | 27.57 | 0.48 |
| O-Haze [50] | Satellite | | Not Specified | | 45 | 45 | 28.96 | 0.80 | 27.49 | 0.37 |
| SOTS [34] | Outdoor | | Not Specified | | 8,964 | 313,950 | 29.27 | **0.99** | 27.44 | 0.83 |
| NH Haze [6] | Outdoor | | Non-Homogenous | | 55 | 55 | 28.43 | 0.66 | 27.70 | 0.22 |
| Haze1k [25] | Satellite | ✗ | ✓ | ✗ | 1,035 | 1,035 | 28.51 | 0.91 | 27.49 | 0.23 |
| RS Haze [50] | Satellite | ✗ | ✓ | ✗ | 6,000 | 54,000 | 27.57 | 0.97 | 27.27 | 0.52 |
| **HazeSpace2M** | Outdoor | ✓ | ✗ | ✓ | 35,936 | **718,720** | 30.91 | 0.98 | 27.11 | 0.25 |
| | Street | ✓ | ✗ | ✓ | 39,998 | 799,960 | 31.91 | 0.98 | 27.36 | 0.39 |
| | **Farmland** | ✓ | ✗ | ✓ | 830 | 16,600 | 32.32 | 0.97 | **27.08** | 0.23 |
| | Satellite | ✗ | ✓ | ✗ | 53,429 | 534,290 | **34.61** | 0.98 | 27.49 | 0.23 |

### 3.2 Annotation Process and Tools

Inspired by [25] and [50], we utilized Adobe Photoshop 25.1 with its advanced ML-based Neural Filters [2] to generate hazy images for our HazeSpace2M dataset. We crafted Photoshop actions [3], which automate editing tasks, to create varied haze levels. This approach efficiently processed our dataset of over two million images generated over months using three computers.

### 3.3 Quantitative Analysis

The HazeSpace2M dataset, as shown in Table 3, includes Fog and EH hazing on its Outdoor (OD), Street (ST), and Farmland (FL) subsets, while the Satellite (SL) subset is treated with Cloud haze, creating subsets designated as ODF, STF, FLF for Fog; ODEH, STEH, FLEH for EH; and SLC for Cloud haze. The OD, ST, FL, and SL subsets consist of synthetic hazy images alongside RHT a subset of real hazy images. Fog and EH haze types applied across ten intensity levels to the OD subset's 35,936 GT images result in 718,720 hazy images for OD, equally split between ODF and ODEH. The ST and FL subsets yield 816,560 hazy images from 40,828 GT images, and SL comprises 534,290 Cloud-hazy images from 53,429 GT images. Totaling around 130,193 GT images, the HazeSpace2M spans approximately 2.2 million hazy images when considering all three haze types and ten haze intensities per GT image, detailed in Table 3.

Compared to established datasets in literature [4–6, 12, 25, 34, 50, 58, 59], Table 4 presents the comparative PSNR and SSIM metrics. The HazeSpace2M dataset demonstrates high PSNR and SSIM at the lowest haze level, reflecting clear images under minimal hazing. At the highest haze level, these metrics show a marked reduction, illustrating the substantial impact of intense hazing. This variance signifies the dataset's wide range of haze intensities, providing a broader scope for analysis than previous datasets. The HazeSpace2M also exceeds others in image volume, offering an extensive array of GT and hazy images. Including the FL subset introduces a new scene type to the dataset, enriching the diversity and research applicability. Comparing the scene types and the haze types, it is evident that the HazeSpace2M consists of a diverse type of scene and haze compared to the existing haze image datasets. Each subset within HazeSpace2M contains GT and hazy images, establishing its superiority in dataset quantity and scene variety.

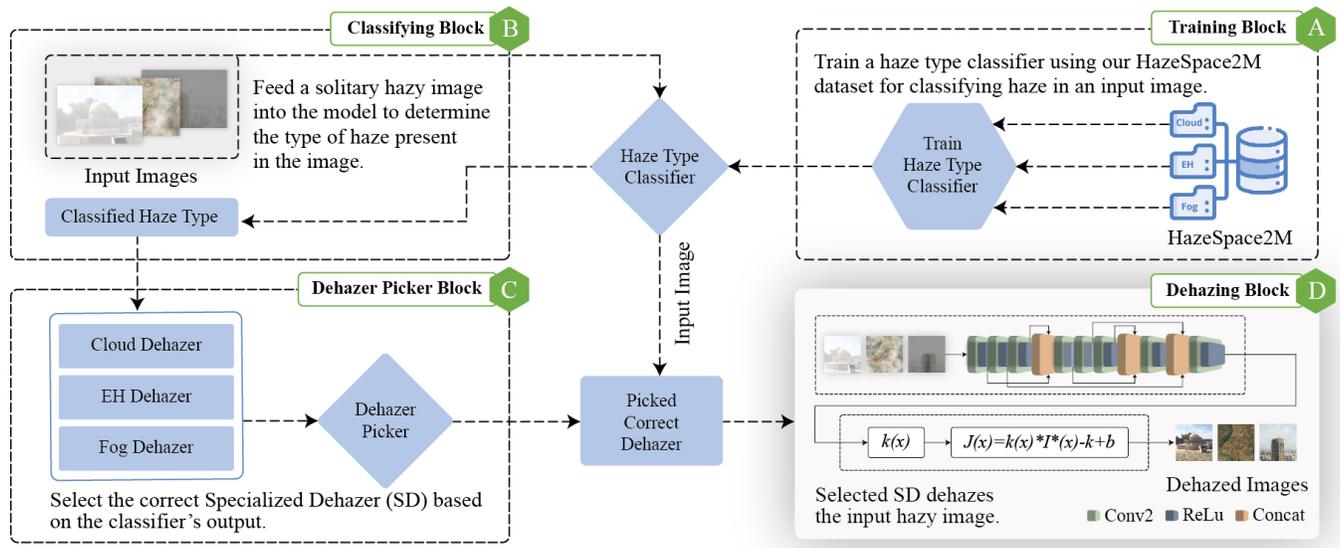

Figure 2: Our proposed framework for specialized dehazer-based intelligent dehazing based on the haze type classification in image enhancement workflows, including (A) training classifiers to recognize haze types, (B) using the classifier to identify the type of haze in a single input image, (C) selecting the appropriate dehazer based on the haze classification, and (D) the final dehazing process to clear the image from atmospheric obscurations with the selected specialized dehazer.

## 4 PROPOSED FRAMEWORK

Our proposed approach for intelligent dehazing based on haze type classification is illustrated in Figure 2. It has four main blocks, each with a particular task. As shown in Figure 2A, we use the dataset to train the SOTA classification models [21, 22, 26, 27, 30, 38, 41, 47, 49, 53–56] to identify which models can classify haze in a single-input image. Thus, we benchmark the SOTA models against existing synthetic hazy benchmarking datasets and the Real Hazy Testset of the HazeSpace2M. Then, as illustrated in Figure 2B, we use the trained classifier for classifying the haze in a single input image. In this paper, we train the SOTA classification models on the HazeSpace2M dataset for haze type classification. As shown in Figure 2C, based on the classification result and the output haze type, the model selects a suitable dehazer and performs the inference accordingly in the Inference Block, which is illustrated in Figure 2D. As with the classification models, we train three dehazing algorithms for three different hazes, namely Fog, EH, and Cloud, on our HazeSpace2M dataset. These dehazers are trained based on the modified ASM [60] over 100 epochs. Utilizing these three specialized dehazers, the complete framework we propose for specialized dehazers-based intelligent dehazing based on haze type classification is depicted in Figure 2.

### 4.1 Experimental Setups

We conduct several experiments in line with the methodology illustrated in Figure 2. Our focus begins with training and evaluating classification models, followed by assessing generalized and specialized dehazers using the HazeSpace2M dataset.

**Haze Type Classification.** For training and validating the classification models, we take subsets from the HazeSpace2M dataset and split them as follows:

*Train and Validation Dataset:* We train our models using 15,000 images from the HazeSpace2M dataset, evenly split into three haze types with 5,000 images each. We allocate 85% (12,750) for training and 15% (2,250) for validation.

*Test Dataset:* We assess models on synthetic and real-life hazy images, creating different sets of testing datasets using the existing Hazy Benchmarking Datasets (HBDs) [5, 12, 25, 34, 50, 58, 59]. We also test the models against the RHT subset of the HazeSpace2M.

To ensure uniform training, all models used a batch size 32, a 0.001 learning rate, and a 512-pixel resolution. Following the footsteps of DTMIC [31], each model underwent a 50-epoch training with a 10-step patience early stopping technique.

**Single Image Dehazing.** We introduce two terms, namely Specialized Dehazer and Generalized Dehazer, and defined below to differentiate between the training processing for each.

*Specialized Dehazer (SD):* This term refers to a dehazing model explicitly trained on images of a particular type of haze. For instance, a model trained exclusively on foggy images to dehaze fog-related obscurities is considered an SD.

*Generalized Dehazer (GD):* In contrast, the Generalized Dehazing model is trained on a broader spectrum of hazy images. The GD model is not limited to a specific type of haze but is designed to handle various hazy conditions.

To conduct experiments with SD and GD, we utilize the same dehazer architecture as depicted in Figure 2D. This architecture is developed based on the modified Atmospheric Scattering Model (ASM) [60] for removing haze in a single input image as follows:

$$I(x) = J(x) \times t(x) + A \times (1 - t(x)). \quad (1)$$

here $K(x)$ represents a combined variable that encapsulates both $t(x)$ and $A$, while $I(x)$ signifies the observed hazy image. $A$ is the global atmospheric light and $t(x)$ is the transmission map as follows:

$$t(x) = e^{-\beta d(x)}, \quad (2)$$

where $\beta$ is the scattering coefficient of the atmosphere, and $d(x)$ is the distance between the object and the camera. The modified version of Eq. (1) that is proposed for LDNet gives improved performance for removing haze from the images, which is verified through comprehensive inferences on different datasets [60]. Hence, for the experiments in our paper, we employ the modified version of the ASM model that is stated as follows:

$$J(x) = K(x) \times I(x) - K(x) + b_{\text{bias}}, \quad (3)$$

here the bias term is incorporated with a default value of 1 and the encapsulated values of $t(x)$ and $A$, which we define by $K(x)$, as:

$$K(x) = \frac{\frac{1}{t(x)} \times (I(x) - A) + (A - b_{\text{bias}})}{(I(x) - 1)}. \quad (4)$$

Table 5: Evaluation of the SOTA classifiers using accuracy (ACC), precision (PRE), and recall (REC) on various combinations of the hazy benchmarking datasets, highlighting their effectiveness for haze type classification after training on the HazeSpace2M.

| Models | Different Combinations of Hazy Benchmarking Datasets for Haze Type Classification ||||||||||||  Average ACC |
|---|---|---|---|---|---|---|---|---|---|---|---|---|---|
| | Fog: FRIDA / EH: O-Haze / Cloud: Haze1k ||| Fog: Cityscapes / EH: NH-Haze / Cloud: Haze1k ||| Fog: Cityscapes / EH: NH-Haze / Cloud: RS-Haze ||| Fog: Cityscapes / EH: O-Haze / Cloud: Haze1k ||| |
| | ACC | PRE | REC | ACC | PRE | REC | ACC | PRE | REC | ACC | PRE | REC | |
| **AlexNet** [30] | 0.96 | 0.96 | 0.96 | 0.95 | 0.95 | 0.95 | 0.83 | 0.91 | 0.83 | 0.96 | 0.96 | 0.96 | 92.50 |
| ConvNextLarge [38] | 0.88 | 0.93 | 0.88 | 0.86 | 0.92 | 0.86 | 0.80 | 0.93 | 0.80 | 0.88 | 0.94 | 0.88 | 85.50 |
| DenseNet121 [26] | **0.98** | 0.98 | 0.98 | 0.90 | 0.96 | 0.90 | 0.63 | 0.95 | 0.63 | 0.90 | 0.97 | 0.90 | 85.25 |
| DenseNet161 [26] | 0.91 | 0.92 | 0.91 | 0.89 | 0.95 | 0.89 | 0.68 | 0.87 | 0.68 | 0.88 | 0.94 | 0.88 | 84.00 |
| DenseNet169 [26] | 0.94 | 0.93 | 0.94 | 0.94 | 0.95 | 0.94 | 0.73 | 0.87 | 0.73 | 0.93 | 0.95 | 0.93 | 88.50 |
| DenseNet201 [26] | 0.96 | 0.96 | 0.96 | **0.96** | 0.96 | 0.96 | 0.78 | 0.94 | 0.78 | **0.97** | 0.97 | 0.97 | 91.75 |
| EfficientNet_B0 [56] | 0.88 | 0.95 | 0.88 | 0.85 | 0.95 | 0.85 | 0.63 | 0.92 | 0.63 | 0.85 | 0.96 | 0.85 | 80.25 |
| EfficientNetV2Large [56] | 0.90 | 0.93 | 0.90 | 0.87 | 0.91 | 0.87 | 0.65 | 0.88 | 0.65 | 0.88 | 0.93 | 0.88 | 82.50 |
| GoogleNet [53] | 0.86 | 0.86 | 0.86 | 0.88 | 0.89 | 0.88 | 0.74 | 0.86 | 0.74 | 0.89 | 0.90 | 0.89 | 84.25 |
| Inception_V3 [54] | 0.78 | 0.86 | 0.78 | 0.79 | 0.89 | 0.79 | 0.68 | 0.90 | 0.68 | 0.80 | 0.91 | 0.80 | 76.25 |
| MNasNet [55] | 0.94 | 0.95 | 0.94 | 0.82 | 0.94 | 0.82 | 0.52 | 0.93 | 0.52 | 0.82 | 0.95 | 0.82 | 77.50 |
| MobileNetV2 [47] | 0.92 | 0.95 | 0.92 | 0.80 | 0.95 | 0.80 | 0.70 | 0.96 | 0.70 | 0.81 | 0.96 | 0.81 | 80.75 |
| MobileNetV3 [22] | 0.76 | 0.94 | 0.76 | 0.51 | 0.92 | 0.51 | 0.43 | 0.95 | 0.43 | 0.51 | 0.94 | 0.51 | 55.25 |
| **ResNet50** [21] | 0.96 | 0.96 | 0.96 | 0.95 | 0.94 | 0.95 | **0.84** | 0.92 | 0.84 | 0.96 | 0.96 | 0.96 | **92.75** |
| ResNet101 [21] | **0.98** | 0.98 | 0.98 | 0.94 | 0.96 | 0.94 | 0.78 | 0.92 | 0.78 | 0.94 | 0.95 | 0.94 | 91.00 |
| ResNet152 [21] | 0.97 | 0.97 | 0.97 | 0.94 | 0.96 | 0.94 | 0.76 | 0.93 | 0.76 | 0.94 | 0.96 | 0.94 | 90.25 |
| ShuffleNetV2 [41] | 0.86 | 0.87 | 0.86 | 0.90 | 0.91 | 0.90 | 0.76 | 0.94 | 0.76 | 0.90 | 0.91 | 0.90 | 85.50 |
| SqueezeNet1 [27] | 0.96 | 0.96 | 0.96 | 0.90 | 0.94 | 0.90 | 0.71 | 0.96 | 0.71 | 0.91 | 0.96 | 0.91 | 87.00 |
| VGG16 [49] | 0.95 | 0.94 | 0.95 | 0.93 | 0.92 | 0.93 | **0.84** | 0.92 | 0.84 | 0.95 | 0.94 | 0.95 | 91.75 |

For the experiments of single image dehazing and to know if SD performs better than GD, we use LDNet [60] that is developed based on Eq. (3) with the same hyperparameter settings as the backbone of our dehazer algorithms and train them with our mentioned training datasets in different steps as follows:

- LDNet: Trained using the RESIDE [34] dataset, a common benchmark in dehazing research.
- GDNet: Trained on 150,000 images, with 50,000 from each category: Fog, Cloud, and EH, for a generalized model.
- SDNets: Individually trained on distinct haze types. Initially, the model is trained exclusively on Fog type hazy images, followed by training on Cloud type, and finally on EH type hazy images, with each model saved after training.

With these setups of the dehazers mentioned above, we organize the training, validation, and test datasets in the following manner:

*Train and Validation Dataset:* For SD models targeting Fog, EH, and Cloud, we use 5,000 GT images and their 50,000 corresponding hazy images at ten intensity levels from each class in the HazeSpace2M dataset. Each SD model is trained with 50,000 hazy images. For the GD model, we combine 50,000 hazy images from each class, creating a 150,000-image dataset. Both datasets are split 90/10 for training and validation.

*Test Dataset:* To evaluate the performance of the SD and GD models for different types of hazes, we curated a test subset from the HazeSpace2M dataset with the images unseen to the model. This subset comprises 1,000 hazy images for each haze category, distributed across ten distinct intensity levels. This test dataset verifies the models' robustness and effectiveness in handling a broad spectrum of haze types and varying levels of haze intensity.

## 5 EXPERIMENTAL RESULTS

Our experiments on haze type classification and single image dehazing showcase the versatility and broad applicability of the HazeSpace2M dataset, with results discussed in subsequent sections.

### 5.1 Results of Haze Type Classification

Our evaluation of SOTA classification models on both synthetic and real hazy images commenced with a training phase of 50 epochs, subsequently assessing performance on the HBDs and RHT datasets are presented in Table 5 and Table 6. Initial results highlighted the challenge within the RHT subset, as most models fell short of achieving 80% accuracy. Nonetheless, ResNet50 surpassed this benchmark, showcasing its potential for single image haze type classification despite being only trained for a short period and training with a subset of the HazeSpace2M dataset.

**Table 6: Evaluation of the SOTA models against the Real Hazy Testset (RHT) of the HazeSpace2M dataset.**

| Models | Accuracy | Precision | Recall |
|---|---|---|---|
| AlexNet [30] | 0.70 | 0.71 | 0.70 |
| ConvNextLarge [38] | 0.63 | 0.72 | 0.63 |
| DenseNet121 [26] | 0.46 | 0.69 | 0.46 |
| DenseNet161 [26] | 0.58 | 0.67 | 0.58 |
| DenseNet169 [26] | 0.56 | 0.65 | 0.56 |
| DenseNet201 [26] | 0.68 | 0.71 | 0.68 |
| EfficientNet_B0 [56] | 0.49 | 0.64 | 0.49 |
| EfficientNetV2Large [56] | 0.48 | 0.67 | 0.48 |
| GoogleNet [53] | 0.66 | 0.68 | 0.66 |
| Inception_V3 [54] | 0.54 | 0.63 | 0.54 |
| MNasNet [55] | 0.45 | 0.68 | 0.45 |
| MobileNetV2 [47] | 0.60 | 0.71 | 0.60 |
| MobileNetV3 [22] | 0.60 | 0.71 | 0.60 |
| **ResNet50 [21]** | **0.80** | **0.78** | **0.80** |
| ResNet101 [21] | 0.70 | 0.72 | 0.70 |
| ResNet152 [21] | 0.63 | 0.71 | 0.63 |
| ShuffleNetV2 [41] | 0.67 | 0.72 | 0.67 |
| SqueezeNet1 [27] | 0.65 | 0.73 | 0.65 |
| VGG16 [49] | 0.70 | 0.69 | 0.70 |

Expanding our investigation, as stated in Table 5, some of the SOTA models namely AlexNet, DenseNet201, ResNet50, ResNet101, ResNet152, and VGG16 give over 90% accuracy on average against the different combinations of HBDs, while models like Inception_V3, MNasNet, MobileNetV3 give 70% accuracy below on average. The bold values for each accuracy (ACC) column represent the top accuracy among all the accuracies while testing the models against the corresponding combinations of the HBDs, while the underlined values represent the second-highest accuracies. The average ACC column shows the average accuracy achieved by each model against the HDBs. Exploring this column, we find that the AlexNet achieves an accuracy of 92.50%, while ResNet50 outperforms the AlexNet with a slightly improved accuracy of 92.75%. Analyzing all these facts, we observed that ResNet50 and AlexNet performed robustly throughout the testing of different combinations of the HBDs.

We further evaluate all the models against the RHT to observe the performance of these models on images affected by the real atmospheric haze. As presented in Table 6, we still found ResNet50 to outperform the other models with an accuracy of 80%, while AlexNet, ResNet101, and VGG16 achieved 70% accuracy.

While several models result in very good accuracy on the synthetic datasets and very low accuracy on the RHT, the challenge lies in classifying haze types on a real hazy image. The ResNet50 shows some robustness by giving 80% accuracy, while the other models failed. The inference results on the RHT images are presented in the Figure 3, showing some of the correctly classified RHT images by the ResNet50 model. Overall, the results show the need to develop robust classification models that can outperform the existing SOTA models in the context of atmospheric haze type classification.

### 5.2 Results of Single Image Dehazing

To investigate the effectiveness of SD models over GD models in single image dehazing, we conducted extensive evaluations using the HazeSpace2M dataset. Our methodology involved training the LDNet [60] and its SD and GD variants across three distinct stages.

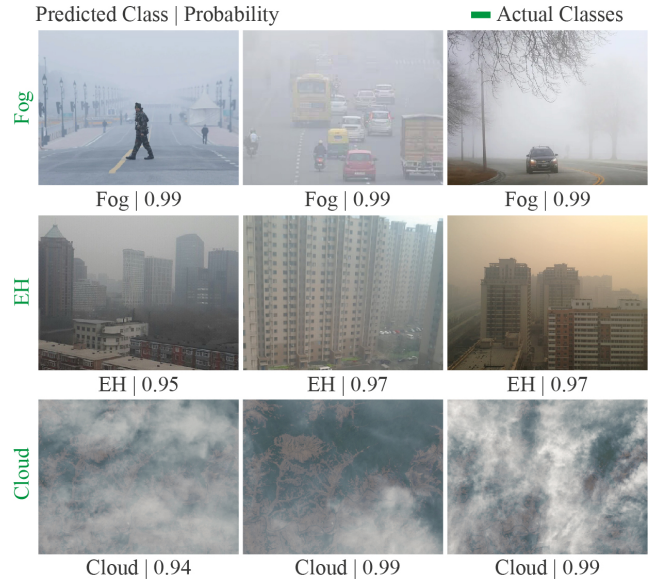

**Figure 3: Hazy images from the RHT showing correct haze type classification by ResNet50 with prediction probabilities.**

Our study rigorously evaluated the original LDNet dehazing model, achieving average PSNR, SSIM, and MSE values of 28.15, 0.65, and 99.89, respectively. We based these averages on comprehensive testing against various haze types, having 1000 hazy images for each haze type with detailed results outlined in Table 7.

Similarly, our evaluation of the GDNet and SDNet models on identical test sets, as detailed in Table 7, reveals that the SD models surpass the LDNet and GDNet in performance. The SDNet models demonstrate PSNR, SSIM, and MSE values of 28.91, 0.82, and 88.36, outperforming the GDNet, which records 28.23, 0.70, and 98.31.

The PSNR values show an improvement of around 2.7% for LDNet compared to SDNets. Similarly, the performance of SDNets improved by 2.4% over GDNet considering the PSNR values from the experimental results presented in Table 7. Moreover, when we examine the SSIM and MSE values, we see a clear performance difference among these models. For example, the SDNets yield around 26.15% higher SSIM and 23.75% improved MSE than the original LDNet. Similarly, compared to the GDNet, the SDNet models show an increase of around 17.14% in SSIM and 10.12% enhancement in MSE for dehazing images affected by Fog, EH, and Cloud. The SDNets outperform the other two models in all three metrics with a PSNR of 28.91, SSIM of 0.82, and MSE of 88.36, showing the effectiveness of an SD model over a GD model.

In addition to comparing performance metrics, the visual examination reveals the superiority of the SD model over the GD models. The single image dehazing examples in Figure 4 demonstrate the visual clarity achieved by the SD model is markedly better than that produced by the original LDNet [60] and GDNet. Figure 4(c) presents the dehazed images of different haze types using the SDNet models. On the other hand, the inference results of the LDNet and GDNet, which we train traditionally with a relatively larger dataset, have been presented in Figure 4(a & b). To compare the visual clarity of the output images from each model, we highlight the differences via the rectangles. Considering PSNR and SSIM, the SDNet models give higher values than both traditional models, where LDNet has trained on RESIDE [34] dataset, and GDNet is trained on a subset of the HazeSpace2M dataset. We ran the inference on the images unknown to the models, i.e., we did not use these images to train the models. It should be noted that the training sets for the SDNet models are 50,000 images, whereas the training set for the GDNet model is 150,000. Even though we train the GDNet model with 3× more different haze types images, SDNets outperform GDNet, ensuring the superiority of our proposed framework.

Table 7: Performance comparison of LDNet, GDNet, and SDNets when subjected to dehazing tasks across various hazy conditions. The average scores reflect the overall performance of each model in processing unknown hazy images.

| Testsets | LDNet [60] | | | GDNet | | | SDNets | | |
|---|---|---|---|---|---|---|---|---|---|
| | PSNR ↑ | SSIM ↑ | MSE ↓ | PSNR ↑ | SSIM ↑ | MSE ↓ | PSNR ↑ | SSIM ↑ | MSE ↓ |
| Fog Testset | 28.47 | 0.78 | 92.46 | 28.47 | 0.77 | 92.31 | **28.55** | **0.85** | **90.49** |
| EH Testset | 27.89 | 0.44 | 105.44 | 27.93 | 0.63 | 104.78 | **28.34** | **0.79** | **98.43** |
| Cloud Testset | 28.11 | 0.75 | 101.76 | 28.29 | 0.70 | 97.85 | **29.84** | **0.83** | **76.17** |
| **Average** | 28.15 | 0.65 | 99.89 | 28.23 | 0.70 | 98.31 | **28.91 (2.41%+)** | **0.82 (17.14%+)** | **88.36 (10.12%+)** |

Table 8: Benchmarking the SOTA Dehazing models.

| Year | SOTA Dehazers | Pre-trained | | | Trained+Proposed Method | | |
|---|---|---|---|---|---|---|---|
| | | PSNR ↑ | SSIM ↑ | MSE ↓ | PSNR ↑ | SSIM ↑ | MSE ↓ |
| 2024 | DEA-Net [10] | 14.02 | 0.7123 | 0.1009 | 34.37 | 0.9447 | 0.0015 |
| 2023 | DehazeFormer [51] | 15.62 | 0.0744 | 0.0744 | 31.32 | 0.9316 | 0.0022 |
| 2023 | C2PNet [72] | 14.01 | 0.7094 | 0.1030 | 30.25 | 0.9048 | 0.0025 |
| 2023 | LHNet [67] | 13.26 | 0.5459 | 99.014 | 25.57 | 0.7568 | 91.644 |
| 2021 | LDNet [60] | 28.15 | 0.6515 | 99.148 | 28.91 | 0.8254 | 88.368 |
| 2021 | DehazeFlow [35] | 25.34 | 0.8581 | 0.0547 | 35.29 | 0.9989 | 0.0010 |
| 2019 | DM2F-Net [14] | 12.35 | 0.5263 | 80.542 | 25.54 | 0.7964 | 0.0037 |

Table 9: Evaluating top SOTA models on foggy datasets.

| Year | Top 3 SOTA Dehazers | FRIDA [58, 59] | | Cityscapes [12] | | HazeSpace2M (Fog) | |
|---|---|---|---|---|---|---|---|
| | | PSNR ↑ | SSIM ↑ | PSNR ↑ | SSIM ↑ | PSNR ↑ | SSIM ↑ |
| 2024 | DEA-Net [10] | 25.14 | 0.8974 | 20.15 | 0.8546 | 14.02 | 0.7123 |
| 2023 | DehazeFormer [51] | 20.65 | 0.1698 | 18.21 | 0.1254 | 15.62 | 0.0744 |
| 2021 | DehazeFlow [35] | 35.54 | 0.9899 | 30.01 | 0.9289 | 25.34 | 0.8581 |

Furthermore, we benchmark the latest SOTA dehazing models in Table 8 to showcase their performance before and after applying our proposed method. For example, the DEA-Net model shows significant improvement in PSNR from 14.02 to 34.37 and SSIM from 0.7123 to 0.9447 after integrating our method, with a notable reduction in MSE from 0.1009 to 0.0015. This trend of enhanced performance with our proposed method is consistent across other models, such as DehazeFormer and C2PNet from 2023, which also exhibit improvements across all metrics. Likewise, we evaluate the top 3 SOTA dehazing models from Table 8 on different foggy datasets, including the fog subset of our HazeSpace2M, and the results are presented in Table 9. The results show that the SOTA models achieve lower PSNR and SSIM when testing on the HazeSpace2M foggy subset than on the FRIDA and Cityscapes datasets, highlighting the challenging nature of our HazeSpace2M dataset.

This data supports the superiority of specialized dehazing techniques. Classifying the type of haze in an image (Figure 2B) and applying the appropriate dehazing method (Figure 2C) enhances model efficacy. Here, our HazeSpace2M leads the way by offering a diverse, large, and challenging hazy dataset with the confirmation of haze type classification, which results in better dehazing.

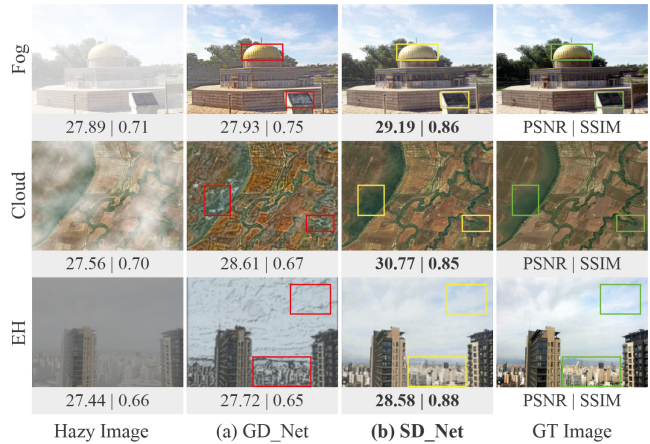

Figure 4: Visual comparison of dehazing results from LDNet, GDNet, and SDNet, with PSNR and SSIM metrics highlighted.

## 6 CONCLUSION

Atmospheric haze poses a significant challenge for autonomous devices, such as self-driving vehicles and drones, that rely on computer vision for navigation. While state-of-the-art dehazing algorithms exist, they are typically trained for a single type of haze, limiting their effectiveness across weather conditions like Fog, Clouds, or EH. Moreover, the need for more diverse datasets hampers the development of robust models. To address this, we introduce HazeSpace2M, a comprehensive dataset of over 2 million images that supports haze type classification and specialized dehazing. While our computational resources are limited for training with the entire dataset, our benchmarks demonstrate that classifying haze types before applying targeted dehazing models significantly improves image clarity, enhancing the performance of autonomous systems and security applications in diverse atmospheric conditions. Future efforts will focus on expanding the dataset's diversity in haze types, depths, and intensities and benchmarking dehazing models to fully leverage HazeSpace2M's potential. Overall, our study demonstrates the significant role of accurate haze type classification in enhancing dehazing outcomes, offering a promising path forward for precision in image processing under adverse weather conditions, thereby filling a crucial gap in the field and setting the stage for future advancements.

# 7 ACKNOWLEDGMENT

This work was supported by the BK21 FOUR Project (Bigdata Research and Education Group for Enhancing Social Connectedness Thorough Advanced Data Technology and Interaction Science Research, 5199990913845), which is funded by the Ministry of Education (MOE, Korea) and the National Research Foundation of Korea (NRF).

## REFERENCES

[1] Abdelrahman Abdelhamed, Mahmoud Afifi, Radu Timofte, and Michael S Brown. 2020. Ntire 2020 challenge on real image denoising: Dataset, methods and results. In *Proceedings of the IEEE/CVF Conference on Computer Vision and Pattern Recognition Workshops*. 496–497.

[2] Adobe. 2024. *Explore a range of creativity with Neural Filters*. Retrieved April 11, 2024 from https://helpx.adobe.com/content/help/en/photoshop/using/neural-filters.html

[3] Adobe. 2024. *Learn how to access various defaults actions available in Photoshop*. Retrieved April 11, 2024 from https://helpx.adobe.com/photoshop/using/actions-actions-panel.html

[4] Cosmin Ancuti, Codruta O Ancuti, Radu Timofte, and Christophe De Vleeschouwer. 2018. I-HAZE: A dehazing benchmark with real hazy and haze-free indoor images. In *Advanced Concepts for Intelligent Vision Systems: 19th International Conference, ACIVS 2018, Poitiers, France, September 24–27, 2018, Proceedings 19*. Springer, 620–631.

[5] Codruta O Ancuti, Cosmin Ancuti, Radu Timofte, and Christophe De Vleeschouwer. 2018. O-haze: a dehazing benchmark with real hazy and haze-free outdoor images. In *Proceedings of the IEEE conference on computer vision and pattern recognition workshops*. 754–762.

[6] Codruta O Ancuti, Cosmin Ancuti, Florin-Alexandru Vasluianu, and Radu Timofte. 2020. Ntire 2020 challenge on nonhomogeneous dehazing. In *Proceedings of the IEEE/CVF Conference on Computer Vision and Pattern Recognition Workshops*. 490–491.

[7] Codruta O Ancuti, Cosmin Ancuti, Florin-Alexandru Vasluianu, Radu Timofte, Han Zhou, Wei Dong, Yangyi Liu, Jun Chen, Huan Liu, Liangyan Li, et al. 2023. NTIRE 2023 HR nonhomogeneous dehazing challenge report. In *Proceedings of the IEEE/CVF Conference on Computer Vision and Pattern Recognition*. 1808–1825.

[8] Tim Brooks, Aleksander Holynski, and Alexei A Efros. 2023. Instructpix2pix: Learning to follow image editing instructions. In *Proceedings of the IEEE/CVF Conference on Computer Vision and Pattern Recognition*. 18392–18402.

[9] BSNV Chaitanya and Snehasis Mukherjee. 2021. Single image dehazing using improved cycleGAN. *Journal of Visual Communication and Image Representation* 74 (2021), 103014.

[10] Zixuan Chen, Zewei He, and Zhe-Ming Lu. 2024. DEA-Net: Single image dehazing based on detail-enhanced convolution and content-guided attention. *IEEE Transactions on Image Processing* (2024).

[11] Marcos V Conde, Gregor Geigle, and Radu Timofte. 2024. High-Quality Image Restoration Following Human Instructions. *arXiv preprint arXiv:2401.16468* (2024).

[12] Marius Cordts, Mohamed Omran, Sebastian Ramos, Timo Rehfeld, Markus Enzweiler, Rodrigo Benenson, Uwe Franke, Stefan Roth, and Bernt Schiele. 2016. The cityscapes dataset for semantic urban scene understanding. In *Proceedings of the IEEE conference on computer vision and pattern recognition*. 3213–3223.

[13] Ilke Demir, Krzysztof Koperski, David Lindenbaum, Guan Pang, Jing Huang, Saikat Basu, Forest Hughes, Devis Tuia, and Ramesh Raskar. 2018. DeepGlobe 2018: A Challenge to Parse the Earth Through Satellite Images. In *The IEEE Conference on Computer Vision and Pattern Recognition (CVPR) Workshops*.

[14] Zijun Deng, Lei Zhu, Xiaowei Hu, Chi-Wing Fu, Xuemiao Xu, Qing Zhang, Jing Qin, and Pheng-Ann Heng. 2019. Deep multi-model fusion for single-image dehazing. In *Proceedings of the IEEE/CVF international conference on computer vision*. 2453–2462.

[15] Vidhi Desai, Sheshang Degadwala, and Dhairya Vyas. 2023. Multi-categories vehicle detection for urban traffic management. In *2023 Second International Conference on Electronics and Renewable Systems (ICEARS)*. IEEE, 1486–1490.

[16] Bosheng Ding, Ruiheng Zhang, Lixin Xu, Guanyu Liu, Shuo Yang, Yumeng Liu, and Qi Zhang. 2023. U 2 D 2 Net: Unsupervised unified image dehazing and denoising network for single hazy image enhancement. *IEEE Transactions on Multimedia* (2023).

[17] Akshay Dudhane, Prashant W Patil, and Subrahmanyam Murala. 2020. An end-to-end network for image de-hazing and beyond. *IEEE Transactions on Emerging Topics in Computational Intelligence* 6, 1 (2020), 159–170.

[18] Masud An-Nur Islam Fahim and Ho Yub Jung. 2020. Single image dehazing using end-to-end deep-dehaze network. In *The 9th International Conference on Smart Media and Applications*. 158–163.

[19] Open Data Canada Government of Canada. 2024. *Forest Fires*. Retrieved April 11, 2024 from https://open.canada.ca/data/en/dataset/9d8f219c-4df0-4481-926f-8a2a532ca003

[20] Jie Gui, Xiaofeng Cong, Yuan Cao, Wenqi Ren, Jun Zhang, Jing Zhang, Jiuxin Cao, and Dacheng Tao. 2023. A comprehensive survey and taxonomy on single image dehazing based on deep learning. *Comput. Surveys* 55, 13s (2023), 1–37.

[21] Kaiming He, Xiangyu Zhang, Shaoqing Ren, and Jian Sun. 2016. Deep residual learning for image recognition. In *Proceedings of the IEEE conference on computer vision and pattern recognition*. 770–778.

[22] Andrew Howard, Mark Sandler, Grace Chu, Liang-Chieh Chen, Bo Chen, Mingxing Tan, Weijun Wang, Yukun Zhu, Ruoming Pang, Vijay Vasudevan, et al. 2019. Searching for mobilenetv3. In *Proceedings of the IEEE/CVF international conference on computer vision*. 1314–1324.

[23] H. Howard, J. McKinley, and A. Elkeurti. 2024. *Wildfire Smoke: Smoky Air Disrupts Life in the Northeast - The New York Times*. Retrieved April 11, 2024 from https://www.nytimes.com/live/2023/06/07/us/canada-wildfires-air-quality-smoke#heres-the-latest-on-the-worsening-air-quality-in-the-us

[24] Cheng-Hsiung Hsieh and Ze-Yu Chen. 2023. Using Haze Level Estimation in Data Cleaning for Supervised Deep Image Dehazing Models. *Electronics* 12, 16 (2023), 3485.

[25] Binghui Huang, Li Zhi, Chao Yang, Fuchun Sun, and Yixu Song. 2020. Single satellite optical imagery dehazing using SAR image prior based on conditional generative adversarial networks. In *Proceedings of the IEEE/CVF winter conference on applications of computer vision*. 1806–1813.

[26] Gao Huang, Zhuang Liu, Laurens Van Der Maaten, and Kilian Q Weinberger. 2017. Densely connected convolutional networks. In *Proceedings of the IEEE conference on computer vision and pattern recognition*. 4700–4708.

[27] Forrest N Iandola, Song Han, Matthew W Moskewicz, Khalid Ashraf, William J Dally, and Kurt Keutzer. 2016. SqueezeNet: AlexNet-level accuracy with 50x fewer parameters and< 0.5 MB model size. *arXiv preprint arXiv:1602.07360* (2016).

[28] RS Jaisurya and Snehasis Mukherjee. 2023. AGLC-GAN: Attention-based global-local cycle-consistent generative adversarial networks for unpaired single image dehazing. *Image and Vision Computing* 140 (2023), 104859.

[29] Zheyan Jin, Shiqi Chen, Yueting Chen, Zhihai Xu, and Huajun Feng. 2023. Let segment anything help image dehaze. (2023). arXiv:arXiv preprint arXiv:2306.15870

[30] Alex Krizhevsky, Ilya Sutskever, and Geoffrey E Hinton. 2012. Imagenet classification with deep convolutional neural networks. *Advances in neural information processing systems* 25 (2012).

[31] Sanjeev Kumar and B Janet. 2022. DTMIC: Deep transfer learning for malware image classification. *Journal of Information Security and Applications* 64 (2022), 103063.

[32] W. Lanzoni and A. L. Matthews. 2024. *What does the sky look like when wildfire smoke is hazardous? | CNN*. Retrieved April 11, 2024 from https://edition.cnn.com/air-quality-wildfire-smoke-pollution-dg/index.html

[33] Boyun Li, Xiao Liu, Peng Hu, Zhongpin Wu, Jiancheng Lv, and Xi Peng. 2022. All-in-one image restoration for unknown corruption. In *Proceedings of the IEEE/CVF Conference on Computer Vision and Pattern Recognition*. 17452–17462.

[34] Boyi Li, Wenqi Ren, Dengpan Fu, Dacheng Tao, Dan Feng, Wenjun Zeng, and Zhangyang Wang. 2018. Benchmarking single-image dehazing and beyond. *IEEE Transactions on Image Processing* 28, 1 (2018), 492–505.

[35] Hongyu Li, Jia Li, Dong Zhao, and Long Xu. 2021. Dehazeflow: Multi-scale conditional flow network for single image dehazing. In *Proceedings of the 29th ACM International Conference on Multimedia*. 2577–2585.

[36] Yuenan Li, Yuhang Liu, Qixin Yan, and Kuangshi Zhang. 2020. Deep dehazing network with latent ensembling architecture and adversarial learning. *IEEE Transactions on Image Processing* 30 (2020), 1354–1368.

[37] Ye Liu, Lei Zhu, Shunda Pei, Huazhu Fu, Jing Qin, Qing Zhang, Liang Wan, and Wei Feng. 2021. From synthetic to real: Image dehazing collaborating with unlabeled real data. In *Proceedings of the 29th ACM international conference on multimedia*. 50–58.

[38] Zhuang Liu, Hanzi Mao, Chao-Yuan Wu, Christoph Feichtenhofer, Trevor Darrell, and Saining Xie. 2022. A convnet for the 2020s. In *Proceedings of the IEEE/CVF conference on computer vision and pattern recognition*. 11976–11986.

[39] Jiaqi Ma, Tianheng Cheng, Guoli Wang, Qian Zhang, Xinggang Wang, and Lefei Zhang. 2023. Prores: Exploring degradation-aware visual prompt for universal image restoration. *arXiv preprint arXiv:2306.13653* (2023).

[40] Kede Ma, Wentao Liu, and Zhou Wang. 2015. Perceptual evaluation of single image dehazing algorithms. In *2015 IEEE International Conference on Image Processing (ICIP)*. IEEE, 3600–3604.

[41] Ningning Ma, Xiangyu Zhang, Hai-Tao Zheng, and Jian Sun. 2018. Shufflenet v2: Practical guidelines for efficient cnn architecture design. In *Proceedings of the European conference on computer vision (ECCV)*. 116–131.

[42] Chippy M Manu and KG Sreeni. 2023. GANID: a novel generative adversarial network for image dehazing. *The Visual Computer* 39, 9 (2023), 3923–3936.

[43] Patrick Morrell. 2024. *Wildfire smoke has led to 'high levels of air pollution' in Toronto, Environment Canada says*. Retrieved April 11, 2024 from https://www.cbc.ca/news/canada/toronto/special-air-quality-statement-toronto-forest-fire-smoke-1.6866738

[44] Ozan Özdenizci and Robert Legenstein. 2023. Restoring vision in adverse weather conditions with patch-based denoising diffusion models. *IEEE Transactions on Pattern Analysis and Machine Intelligence* (2023).
[45] Dongwon Park, Byung Hyun Lee, and Se Young Chun. 2023. All-in-one image restoration for unknown degradations using adaptive discriminative filters for specific degradations. In *2023 IEEE/CVF Conference on Computer Vision and Pattern Recognition (CVPR)*. IEEE, 5815–5824.
[46] Vaishnav Potlapalli, Syed Waqas Zamir, Salman Khan, and Fahad Shahbaz Khan. 2023. Promptir: Prompting for all-in-one blind image restoration. *arXiv preprint arXiv:2306.13090* (2023).
[47] Mark Sandler, Andrew Howard, Menglong Zhu, Andrey Zhmoginov, and Liang-Chieh Chen. 2018. Mobilenetv2: Inverted residuals and linear bottlenecks. In *Proceedings of the IEEE conference on computer vision and pattern recognition*. 4510–4520.
[48] Gege Shi, Xueyang Fu, Chengzhi Cao, and Zheng-Jun Zha. 2023. Alleviating Spatial Misalignment and Motion Interference for UAV-based Video Recognition. In *Proceedings of the 31st ACM International Conference on Multimedia*. 193–202.
[49] Karen Simonyan and Andrew Zisserman. 2015. Very deep convolutional networks for large-scale image recognition. In *3rd International Conference on Learning Representations (ICLR 2015)*.
[50] Yuda Song, Zhuqing He, Hui Qian, and Xin Du. 2023. Vision transformers for single image dehazing. *IEEE Transactions on Image Processing* 32 (2023), 1927–1941.
[51] Yuda Song, Zhuqing He, Hui Qian, and Xin Du. 2023. Vision transformers for single image dehazing. *IEEE Transactions on Image Processing* 32 (2023), 1927–1941.
[52] Ziyi Sun, Yunfeng Zhang, Fangxun Bao, Kai Shao, Xinxin Liu, and Caiming Zhang. 2021. ICycleGAN: Single image dehazing based on iterative dehazing model and CycleGAN. *Computer Vision and Image Understanding* 203 (2021), 103133.
[53] Christian Szegedy, Wei Liu, Yangqing Jia, Pierre Sermanet, Scott Reed, Dragomir Anguelov, Dumitru Erhan, Vincent Vanhoucke, and Andrew Rabinovich. 2015. Going deeper with convolutions. In *Proceedings of the IEEE conference on computer vision and pattern recognition*. 1–9.
[54] Christian Szegedy, Vincent Vanhoucke, Sergey Ioffe, Jon Shlens, and Zbigniew Wojna. 2016. Rethinking the inception architecture for computer vision. In *Proceedings of the IEEE conference on computer vision and pattern recognition*. 2818–2826.
[55] Mingxing Tan, Bo Chen, Ruoming Pang, Vijay Vasudevan, Mark Sandler, Andrew Howard, and Quoc V Le. 2019. Mnasnet: Platform-aware neural architecture search for mobile. In *Proceedings of the IEEE/CVF conference on computer vision and pattern recognition*. 2820–2828.
[56] Mingxing Tan and Quoc Le. 2019. Efficientnet: Rethinking model scaling for convolutional neural networks. In *International conference on machine learning*. PMLR, 6105–6114.
[57] Yu Xiang Tan, Malika Meghjani, and Marcel Bartholomeus Prasetyo. 2023. Localization with Anticipation for Autonomous Urban Driving in Rain. (2023). arXiv:arXiv:2306.09134
[58] Jean-Philippe Tarel, Nicholas Hautiere, Laurent Caraffa, Aurélien Cord, Houssam Halmaoui, and Dominique Gruyer. 2012. Vision enhancement in homogeneous and heterogeneous fog. *IEEE Intelligent Transportation Systems Magazine* 4, 2 (2012), 6–20.
[59] Jean-Philippe Tarel, Nicolas Hautiere, Aurélien Cord, Dominique Gruyer, and Houssam Halmaoui. 2010. Improved visibility of road scene images under heterogeneous fog. In *2010 IEEE intelligent vehicles symposium*. IEEE, 478–485.
[60] Hayat Ullah, Khan Muhammad, Muhammad Irfan, Saeed Anwar, Muhammad Sajjad, Ali Shariq Imran, and Victor Hugo C de Albuquerque. 2021. Light-DehazeNet: a novel lightweight CNN architecture for single image dehazing. *IEEE transactions on image processing* 30 (2021), 8968–8982.
[61] U.S. Department of Transportation, Federal Highway Adminstration. 2023. *Low Visibility*. Retrieved April 11, 2024 from https://ops.fhwa.dot.gov/weather/weather_events/low_visibility.htm
[62] Jeya Maria Jose Valanarasu, Rajeev Yasarla, and Vishal M Patel. 2022. Transweather: Transformer-based restoration of images degraded by adverse weather conditions. In *Proceedings of the IEEE/CVF Conference on Computer Vision and Pattern Recognition*. 2353–2363.
[63] Tao Wang, Guangpin Tao, Wanglong Lu, Kaihao Zhang, Wenhan Luo, Xiaoqin Zhang, and Tong Lu. 2024. Restoring vision in hazy weather with hierarchical contrastive learning. *Pattern Recognition* 145 (2024), 109956.
[64] Xintao Wang, Ke Yu, Chao Dong, and Chen Change Loy. 2018. Recovering realistic texture in image super-resolution by deep spatial feature transform. In *Proceedings of the IEEE conference on computer vision and pattern recognition*. 606–615.
[65] Yan Yang, Jinlong Zhang, Zhiwei Wang, and Haowen Zhang. 2023. Single image fast dehazing based on haze density classification prior. *Expert Systems with Applications* 232 (2023), 120777.
[66] Mingde Yao, Ruikang Xu, Yuanshen Guan, Jie Huang, and Zhiwei Xiong. 2023. Neural Degradation Representation Learning for All-In-One Image Restoration. *arXiv preprint arXiv:2310.12848* (2023).
[67] Shenghai Yuan, Jijia Chen, Jiaqi Li, Wenchao Jiang, and Song Guo. 2023. LHNet: A Low-cost Hybrid Network for Single Image Dehazing. In *Proceedings of the 31st ACM International Conference on Multimedia*. 7706–7717.
[68] Amir Roshan Zamir and Mubarak Shah. 2014. Image geo-localization based on multiplenearest neighbor feature matching usinggeneralized graphs. *IEEE transactions on pattern analysis and machine intelligence* 36, 8 (2014), 1546–1558.
[69] Cheng Zhang, Yu Zhu, Qingsen Yan, Jinqiu Sun, and Yanning Zhang. 2023. All-in-one multi-degradation image restoration network via hierarchical degradation representation. In *Proceedings of the 31st ACM International Conference on Multimedia*. 2285–2293.
[70] Yuxiao Zhang, Alexander Carballo, Hanting Yang, and Kazuya Takeda. 2023. Perception and sensing for autonomous vehicles under adverse weather conditions: A survey. *ISPRS Journal of Photogrammetry and Remote Sensing* 196 (2023), 146–177.
[71] Yuan Zhang, Youpeng Sun, Zheng Wang, and Ying Jiang. 2023. YOLOv7-RAR for urban vehicle detection. *Sensors* 23, 4 (2023), 1801.
[72] Yu Zheng, Jiahui Zhan, Shengfeng He, Junyu Dong, and Yong Du. 2023. Curricular contrastive regularization for physics-aware single image dehazing. In *Proceedings of the IEEE/CVF conference on computer vision and pattern recognition*. 5785–5794.
[73] Bolei Zhou, Hang Zhao, Xavier Puig, Sanja Fidler, Adela Barriuso, and Antonio Torralba. 2017. Scene parsing through ade20k dataset. In *Proceedings of the IEEE conference on computer vision and pattern recognition*. 633–641.
[74] Bolei Zhou, Hang Zhao, Xavier Puig, Tete Xiao, Sanja Fidler, Adela Barriuso, and Antonio Torralba. 2019. Semantic understanding of scenes through the ade20k dataset. *International Journal of Computer Vision* 127, 3 (2019), 302–321.

# (Supplementary Materials)
# HazeSpace2M: A Dataset for Haze Aware Single Image Dehazing

Md Tanvir Islam  
Sungkyunkwan University  
Suwon, Republic of Korea  
tanvirnwu@g.skku.edu

Nasir Rahim  
Sungkyunkwan University  
Suwon, Republic of Korea  
nrahim3797@skku.edu

Saeed Anwar  
The Australian National University  
Canberra, Australia  
saeed.anwar@anu.edu.au

Muhammad Saqib  
National Collections & Marine Infrastructure, CSIRO, Marsfield Sydney, Australia  
muhammad.saqib@data61.csiro.au

Sambit Bakshi  
National Institute of Technology  
Rourkela, India  
bakshisambit@ieee.org

Khan Muhammad*  
Department of Human-AI Interaction  
Sungkyunkwan University  
Seoul, Republic of Korea  
khanmuhammad@g.skku.edu

## 1 QUALITITATIVE ANALYSIS

We analyzed the haze distributions across images and the 10 levels in our HazeSpace2M. As shown in Figure 1, the distributions of the images are varied across different levels of the hazes, which is crucial to generalize the dehazing models on unknown data.

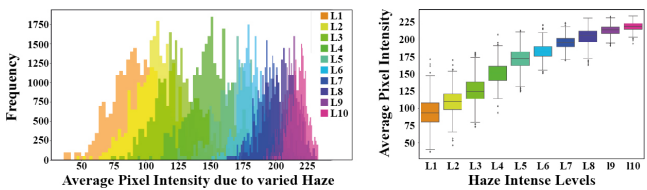

Figure 1: Haze distribution of HazeSpace2M dataset.

We assured the quality of the GT images while picking them from various sources, as also mentioned in Section 3.1 (Data Collection and Generation) of the paper. For example, we eliminated the images from the SOTS [2] dataset, finding the images with hazes present in them. Figure 2 shows some hazy images that we found during data collection, which were disqualified to be the GT images of the HazeSpace2M dataset.

Figure 3 offers a visual range of haze intensities within the HazeSpace2M dataset, showcasing images that progressively intensify in the haze. This array of I mages vividly demonstrates the range of visibility reduction across various environmental conditions, including outdoor scenes, streets, farmlands, and satellite views. Each row corresponds to a different subset, with the transition from left to right depicting a gradual increase from clear to heavily hazed images. The gradation serves as a crucial reference for developing and testing dehazing algorithms, enabling a nuanced understanding of how different haze levels affect image perception. This representation underscores the dataset's versatility and richness, making it a valuable resource for researchers aiming to improve image clarity in diverse atmospheric conditions. The visual gradation also highlights the dataset's potential to train models that can accurately classify haze types. This ensures optimal image enhancement in a wide range of real-world multimedia applications.

*Corresponding author

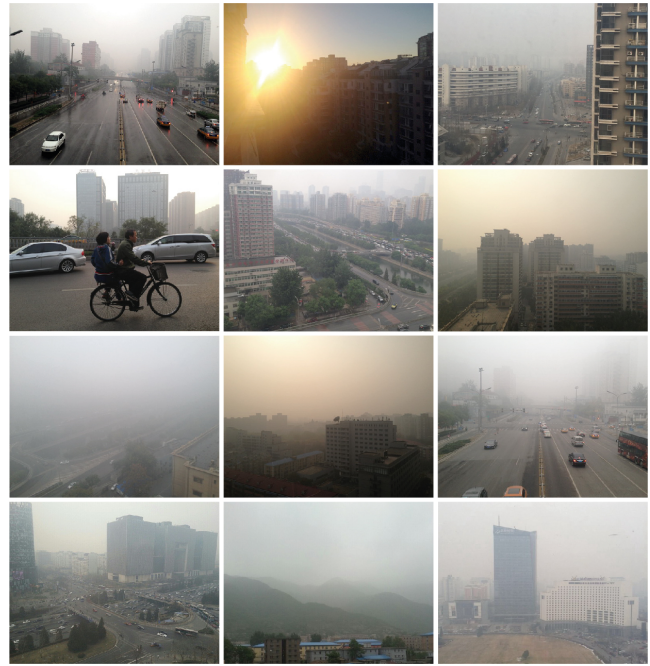

Figure 2: Sample images that we eliminated while collecting images for use as the GT images in the HazeSpace2M dataset.

We found the ResNet50 [1] model to give the best accuracy, as discussed in Section 5.1 (Results of Haze Type Classification) when evaluating against the synthetic benchmarking datasets and the real hazy image dataset for classifying the haze type presented in a single input image. Figure 4(a) reveals the training and validation loss curves for the ResNet50 [1] model. These curves illustrate a decrease in validation loss relative to training loss over time. Additionally, the figure indicates that the model's training was halted after 35 epochs using the early stopping technique. The confusion matrix for the same model is also presented in Figure 4(b), which shows that even though the model classifies the Cloud-type haze with higher accuracy, it struggles to classify the two other hazes, namely Fog and EH. The confusion matrix and the precision-recall curve are presented in Figure 4(b & c), showing that some of the hazes were classified as EH while they were actually Fog and vice versa. This is because of the nature of the hazes. In reality, the EH and Fog-type hazes are quite similar, validating the challenges in classifying haze types in real images.

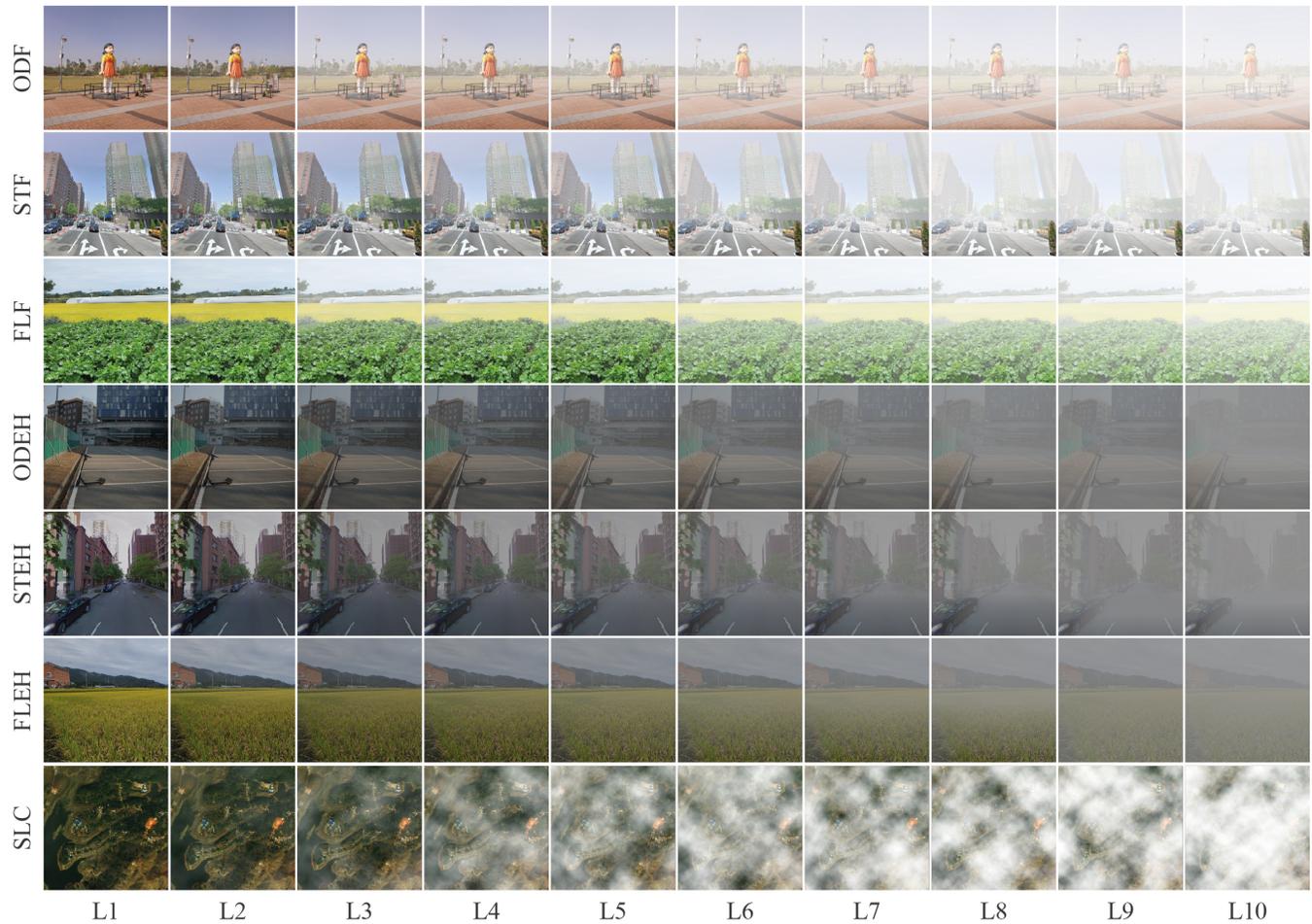

Figure 3: Examples of haze intensities across different scene types within the HazeSpace2M dataset. Each row represents a subset (Outdoor, Street, Farmland, and Satellite) with images transitioning from low to high haze levels, following the order of left to right. To avoid repetition, abbreviations used in this figure are not defined and are given in Table 3 of the main paper.

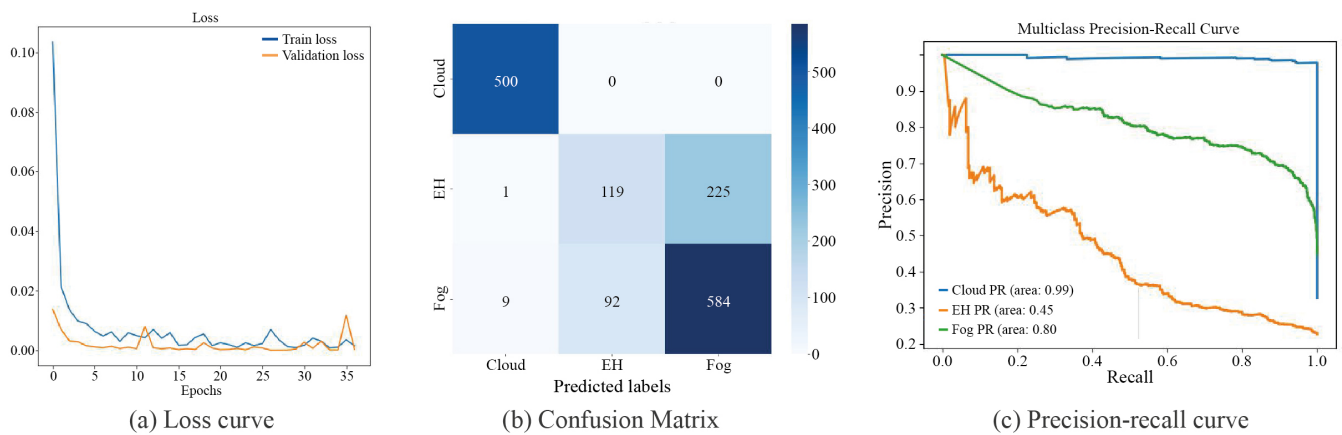

Figure 4: (a) Train and validation loss curve for training the ResNet50 model, (b) Confusion matrix of ResNet50, and (c) Precision-recall curve of the same model while evaluating it on the Real Hazy Testset (RHT) of the HazeSpace2M dataset. ResNet50 is selected as a sample because its result is the highest on the RHT.

Table 1: Comparative analysis of PSNR across varying haze levels. This table details the PSNR values for different haze intensities (L1 to L10) within the HazeSpace2M dataset, divided by scene types (Outdoor, Street, Farmland, Satellite) and haze types (Fog, EH, Cloud). These metrics provide insights into the consistency of image quality amidst diverse hazy environments.

| Scene | Haze Type | Average PSNR for different levels of haze | | | | | | | | | |
|---|---|---|---|---|---|---|---|---|---|---|---|
| | | L1 | L2 | L3 | L4 | L5 | L6 | L7 | L8 | L9 | L10 |
| Outdoor | Fog | 29.32 | 28.00 | 27.70 | 27.70 | 27.77 | 27.71 | 27.72 | 27.71 | 27.74 | 27.78 |
| | EH | 27.74 | 27.85 | 27.86 | 27.86 | 27.91 | 27.89 | 27.89 | 27.87 | 27.82 | 27.87 |
| Street | Fog | 30.53 | 29.49 | 28.95 | 28.63 | 28.45 | 28.37 | 28.20 | 28.20 | 28.16 | 28.17 |
| | EH | 27.68 | 27.72 | 27.86 | 27.92 | 27.84 | 27.78 | 27.76 | 27.82 | 27.87 | 27.79 |
| Farmland | Fog | 29.67 | 28.26 | 27.82 | 28.26 | 27.82 | 27.80 | 27.75 | 27.76 | 27.80 | 27.85 |
| | EH | 27.70 | 27.77 | 27.87 | 27.92 | 27.96 | 27.92 | 27.89 | 27.88 | 27.82 | 27.80 |
| Satellite | Cloud | 29.67 | 29.49 | 27.58 | 27.49 | 27.63 | 27.72 | 27.87 | 27.99 | 28.07 | 28.11 |
| | **Average** | **28.90** | **28.37** | **27.95** | **27.97** | **27.91** | **27.88** | **27.86** | **27.89** | **27.89** | **27.91** |

Table 2: Evaluation of image quality across haze intensity levels. This table shows average SSIM scores for different levels of haze (L1 to L10) within the HazeSpace2M dataset across various scenes (Outdoor, Street, Farmland, Satellite) and corresponding haze types (Fog, EH, Cloud), illustrating the dataset's utility for image quality assessment under varying hazy conditions.

| Scene | Haze Type | Average SSIM for different levels of haze | | | | | | | | | |
|---|---|---|---|---|---|---|---|---|---|---|---|
| | | L1 | L2 | L3 | L4 | L5 | L6 | L7 | L8 | L9 | L10 |
| Outdoor | Fog | 0.97 | 0.94 | 0.83 | 0.83 | 0.79 | 0.71 | 0.68 | 0.66 | 0.64 | 0.60 |
| | EH | 0.96 | 0.91 | 0.83 | 0.81 | 0.77 | 0.73 | 0.70 | 0.65 | 0.63 | 0.59 |
| Street | Fog | 0.98 | 0.97 | 0.94 | 0.90 | 0.86 | 0.79 | 0.74 | 0.74 | 0.67 | 0.63 |
| | EH | 0.96 | 0.91 | 0.84 | 0.82 | 0.78 | 0.74 | 0.70 | 0.66 | 0.65 | 0.60 |
| Farmland | Fog | 0.97 | 0.94 | 0.56 | 0.94 | 0.56 | 0.51 | 0.67 | 0.64 | 0.61 | 0.57 |
| | EH | 0.94 | 0.88 | 0.79 | 0.76 | 0.72 | 0.68 | 0.64 | 0.61 | 0.56 | 0.51 |
| Satellite | Cloud | 0.97 | 0.98 | 0.94 | 0.85 | 0.76 | 0.72 | 0.66 | 0.60 | 0.56 | 0.52 |
| | **Average** | **0.96** | **0.93** | **0.82** | **0.84** | **0.75** | **0.70** | **0.68** | **0.67** | **0.62** | **0.57** |

## 2 PSEUDOCODE

**Pseudocode of the Proposed Framework.** In this paper, we propose a novel technique of specialized dehazers-based smart image dehazing based on the haze type classification. Algorithm 1 presents a step-by-step pseudocode of our proposed framework for processing a hazy image $I_h$ to output the haze type $h_t$ and the corresponding dehazed image $J(x)$. The algorithm requires a dataset $\Omega$, a classifier $C$, and a set of trained dehazers $\Delta\theta$. The process begins by preparing the data, training a classifier on the dataset, classifying the type of haze present in the input image, selecting an appropriate dehazer based on this classification, and finally applying the selected dehazer to produce a clear image.

Table 3 defines the symbols used in the algorithm, connecting the abstract symbols to their concrete meanings within the context of image dehazing. The table serves as a quick reference for understanding the variables and entities involved in the algorithm.

Together, the pseudocode and the table of notations provide a comprehensive overview of the proposed dehazing technique.

Table 3: Descriptions of the notations used in Algorithm 1.

| Symbols | Description | Symbols | Description |
|---|---|---|---|
| $\Omega$ | Dataset | $\tau$ | Train dataset |
| $\nu$ | Validation dataset | $\rho$ | Test dataset |
| $\phi$ | Dehazers | $\Delta\theta$ | Trained dehazers |
| $C$ | Classifier | $h_t$ | Haze type |
| $I_h$ | Hazy input image | $J(x)$ | Dehazed image |
| $\delta$ | Modified ASM model | | |

## 3 QUANTITATIVE ANALYSIS

As it is an extensive dataset, it is highly time-consuming to calculate the Peak Signal-to-Noise Ratio (PSNR) and Structural Similarity Index (SSIM) values for each hazy image. So, we separated 1,000 hazy images from level 1 to level 10 haze intensity as a sample and calculated the average PSNR and SSIM values for each subset and level. Table 1 and Table 2 compare image quality metrics, PSNR, and SSIM across different haze levels within the HazeSpace2M Dataset.

**Average PSNR.** Table 1 presents a comparative analysis of PSNR values across different haze intensities, offering an in-depth view of the image quality within the HazeSpace2M dataset. From left to right of the table, the haze intensity increases as the level increases, which is also evident by the average PSNR values of the images, considered here as a sample. It also breaks down the dataset by scene types (Outdoor, Street, Farmland, Satellite) and haze types (Fog, EH, Cloud), showing the impact of various haze levels on image clarity. This table is a testament to the dataset's comprehensive nature and applicability in assessing image quality in hazy conditions.

**Average SSIM.** Table 2 further complements this by providing average SSIM scores for the same levels of haze intensity, scene, and haze types. SSIM scores provide insight into the perceived quality of images, highlighting the dataset's utility for more subjective assessments of image quality under varying hazy conditions. These metrics collectively underscore the dataset's robustness for developing advanced dehazing algorithms.

## REFERENCES

[1] Kaiming He, Xiangyu Zhang, Shaoqing Ren, and Jian Sun. 2016. Deep residual learning for image recognition. In *Proceedings of the IEEE conference on computer vision and pattern recognition*. 770–778.
[2] Boyi Li, Wenqi Ren, Dengpan Fu, Dacheng Tao, Dan Feng, Wenjun Zeng, and Zhangyang Wang. 2018. Benchmarking single-image dehazing and beyond. *IEEE Transactions on Image Processing* 28, 1 (2018), 492–505.

**Algorithm 1** Pseudocode of our proposed framework for haze aware single image dehazing.

1: **Input:** Hazy image $I_h$
2: **Outputs:** Haze type $h_t$, Dehazed image $J(x)$
3: **Require:** $\Omega$, $C$, $\Delta\theta = \{\phi_{\text{cloud}}, \phi_{\text{EH}}, \phi_{\text{fog}}\}$
4: Data preparation: $[\tau, \nu, \rho] = \text{split\_dataset}(\Omega)$

5: **Procedure:** TrainClassifier($\tau$)
6:     Train classifier $C$ using dataset $\tau$
7:     **return** $C$
8: **End Procedure**

9: **Procedure:** ClassifyHazeType($I_h$, $C$)
10:     Haze type $h_t \leftarrow C(I_h)$
11:     **return** $h_t$
12: **End Procedure**

13: **Procedure:** PickDehazer($h_t$, $\Delta\theta$)
14:     Dehazer $\delta \leftarrow$ select a dehazer from $\Delta\theta$ based on $h_t$
15:     **return** $\delta$
16: **End Procedure**

17: **Procedure:** DehazeImage($I_h$, $\delta$)
18:     Dehazed image $J(x) \leftarrow \delta(I_h)$
19:     **return** $J(x)$
20: **End Procedure**